\documentclass[a4paper,fleqn]{cas-dc}

\usepackage{float}
\usepackage[numbers]{natbib}
\def\tsc#1{\csdef{#1}{\textsc{\lowercase{#1}}\xspace}}
\tsc{WGM}
\tsc{QE}

\begin{document}
\let\WriteBookmarks\relax
\def\floatpagepagefraction{1}
\def\textpagefraction{.001}

\shorttitle{}    


\title [mode = title]{MDA framework: multi-architecture domain adaptation  for unsupervised person re-identification}  

\tnotemark[1] 


%

\author[1]{Wei Zhang}


\fnmark[1]




\affiliation[1]{organization={School of Electronics and Information Technology, Sun Yat-sen University},
            city={Guangzhou},
            state={Guangdong},
            country={China}}

\author[2]{Peijun Ye}

\fnmark[2]




\affiliation[2]{organization={Peng Cheng Laboratory},
            city={Shenzhen},
            state={Guangdong},
            country={China}}
\author[3]{Dihu Chen}

\fnmark[3]




\author[1]{Tao Su}



\cormark[1]

\affiliation[3]{organization={School of Integrated Circuits, Sun Yat-sen University},
	city={Shenzhen},
	state={Guangdong},
	country={China}}



\begin{abstract}
Unsupervised person re-identification not only faces the challenge of domain gaps, but existing methods also rarely support multiple network architectures. To tackle both issues, this paper proposes  a multi-architecture auto-adaptive domain adaptation method: Mamba augmented multi-architecture domain adaptation framework. Specifically, this paper constructs the multi-architecture domain adaptation capability at two levels. At the first level, we propose a composite operator(MDA operator) that encapsulates Mamba as a component for enhancing domain adaptation. At the second level, based on this operator, we propose a Mamba-enhanced multi-architecture Domain Adaptation framework(MDA framework). Comprehensive experiments on person re-identification benchmarks validate the consistency and flexibility of the proposed framework. The MDA framework supports versatile deployment strategies without being confined to a single integration point. Under a single-deployment configuration, it surpasses several SOTA methods. Moreover, it demonstrates substantial performance improvements across both CNN and Transformer architectures. When embedded into a CNN backbone, the framework achieves a  10.2\% mAP improvement on the adaptation task from CUHK03 to Market-1501. Integrated with a Transformer backbone, it yields a 19.8\% mAP improvement on the same task.
\end{abstract}



\maketitle

\section{Introduction}
As a research topic related to security and intelligence, person re-identification (person Re-ID) has long been a popular research direction. Person re-identification is a sub-task of person retrieval. As a branch of person re-identification, unsupervised domain adaptation person re-identification (UDA person Re-ID) aims to transfer the recognition capability from a supervised source domain to an unsupervised target domain. Owing to its wide application in fields such as intelligent person search and smart cities, an increasing number of researchers are becoming engaged in this field.

The domain gap in person Re-ID is primarily manifested by variations in pedestrian appearance across different datasets. This domain divergence is chiefly demonstrated through: discrepancies arising from changes in lighting conditions; differences caused by viewing angles; variations induced by background clutter; and disparities stemming from camera characteristics. Consequently, the robustness of neural network-extracted pedestrian features across domains is crucial for enhancing domain adaptation.

In the realm of person Re-ID, mainstream domain adaptation approaches, while making significant contributions, exhibit several limitations requiring further investigation. These methodologies encompass: feature alignment techniques, cross-domain image style translation approaches, and clustering-based methods. Regarding feature alignment techniques, such methods frequently necessitate additional annotated data to effectively bridge domain gaps. As for cross-domain image style translation, these approaches often present persistent challenges in training stability and pose significant training difficulties. Clustering-based methods commonly contend with the persistent challenge of imbalanced clustered data distribution.

Based on analysis of the Mamba architecture, we observe the potential to leverage Mamba's distinctive characteristics for enhancing domain adaptation. In various computer vision domains, Mamba\cite{gu2023mamba} and its derivative architectures are emerging as solutions to address the spatial inefficiency inherent in self-attention mechanisms, effectively replacing Transformer architecture. The Mamba architecture possesses the following operational property: it selectively propagates or discards information along the sequence based on the current token. The strategic utilization of this property holds significant promise for improving domain adaptation performance.

To address the shortcomings of mainstream methods, we propose a simple yet efficient method from the perspective of enhancing domain adaptation with Mamba\cite{gu2023mamba}. Under domain adaptation settings, our method operates by selectively attending to identity-relevant tokens(feature points)  while disregarding identity-irrelevant ones, thereby improving recognition performance. This approach not only offers a novel alternative for domain adaptation but also overcomes limitations of current methods, as evidenced by: (1) Efficient Training and Convergence. The method demonstrates ease of training and rapid convergence. (2)Minimal Information Requirements. It eliminates the need for auxiliary information from either the source or target domains. (3) Enhanced Clustering Stability. The training process inherently fosters stability in feature clustering. Remarkably, this methodology can even outperform some SOTA domain adaptation solutions.

In summary, the contributions of this article are as follows:
\begin{itemize}
	\item Composite Domain Adaptation Operator. This work proposes a Mamba augmented Domain Adaptation operator (MDA operator) as a foundational building block for domain adaptation.
	\item Mamba augmented multi-architecture domain adaptation framework (MDA framework). Following a scalable design methodology, we construct a domain adaptation framework based on the aforementioned composite operator.
	\item Extensive Experimental Validation. Comprehensive experiments on three Re-ID benchmarks demonstrate that the MDA framework exhibits robust positional independence and enhances domain adaptation across low-, medium-, and high-level feature stages. The method proves effective across multiple baselines and outperforms several SOTA approaches.
	\item Architectural Compatibility: The MDA framework demonstrates compatibility with both Transformer and CNN-based backbone networks, enabling seamless integration and switching between architectures.	
\end{itemize}  

\section{Related Work}

\subsection{domain adaptation techniques for person Re-ID}
Feature-level alignment methods reduce the domain gap by aligning features between the training and inference domains. During the unsupervised domain adaptation process, HUDA\cite{LAN2022108514} jointly modeled both global distribution alignment and local instance alignment to discover transferable knowledge from the source domain. Employing a dual-branch network with an identity inference space, TJ-AIDL\cite{wang2018transferable} leveraged not only the label information but also the attribute annotations of source domain images. This facilitated improved transfer of knowledge learned from the source domain to the target domain. Such approaches frequently rely on auxiliary information or additional annotations. This reliance often incurs additional costs.

Methods for cross-domain image style transfer typically employ various variants of Generative Adversarial Network\cite{10.1145/3422622} to migrate image styles across domains, generating images that possess the style of the target domain while preserving the labels from the source domain. PSI-StarGAN\cite{dungperson} introduced semantic information of humans into the image conversion process to enhance domain adaptation performance. From the perspective of reducing sample annotation costs,  PTGAN\cite{wei2018person} generated images preserving the identity from the source domain with backgrounds from the target domain, to bridge the domain gap. Such approaches generally entail higher training difficulty and longer training durations compared to conventional Convolutional Neural Networks, and the quality of synthesized images is constrained by the GAN’s fitting capability.

Clustering-based approaches employ pre-trained models to extract pedestrian features while estimating pseudo-labels based on clustering results. This model is then fine-tuned on the target domain using images assigned these pseudo-labels.  The  MLC framework \cite{LI2022108521} leveraged a multi-scale network to generate multi-granularity pedestrian features from both global and local perspectives. Subsequently, a multi-label learning module utilized a memory feature bank to assign a multi-label vector to each image based on its similarity to features within the bank. Finally, after several iterations of multi-label training, self-paced clustering  capable of filtering out noisy samples  is incorporated into the training process to assign a pseudo-label to each image. The CSA method\cite{guo2025camera} incorporated a dual cluster consistency refinement mechanism designed to improve the quality of the clustered pseudo-labels. This category of methods is inherently limited by the performance ceiling of the clustering algorithm and susceptibility to label noise, frequently resulting in intra-cluster data imbalance. 

Our approach differs distinctively from the three aforementioned technical pathways while effectively mitigating certain limitations inherent in them. Furthermore, our method does not require any auxiliary information, whether sourced from the source or target domains.  Notably, our solution is characterized by ease of training, rapid convergence, and flexible deployment.

\subsection{Mamba in Re-ID}
Re-identification has now incorporated Mamba. MambaPro\cite{wang2025mambapro} introduced a Mamba aggregation block to model interactions between long sequences from different modalities, enhancing performance in multi-modal object re-identification. MambaReID\cite{zhang2024mambareid}  applied Mamba to multi-modal object re-identification, deploying visual Mamba and dense Mamba to extract multi-modal features. The CLIMB-ReID framework\cite{yu2025climb} proposed a multi-temporal Mamba module for extracting multi-granular spatiotemporal information. While previous Mamba-related research has primarily focused on supervised learning, we extend its application to unsupervised learning, further broadening the scope of Mamba.

\section{Proposed Methodology}
\subsection{Preliminary}
State space models(SSMs) have been widely applied to various sequential data modeling tasks. Inspired by continuous systems, SSMs are adept at capturing dynamic patterns within long sequences. Specifically, a state space model can map an input sequence $X(t) \in \mathbb{R}^{L}$ to an output sequence  $Y(t) \in \mathbb{R}^{L}$, represented by the following expression:

\begin{equation}
	\begin{cases}
		$h'(t) = A*h(t) + B*X(t) $   \\
		$Y(t) = C*h(t) + D*X(t)  $   
	\end{cases}
\end{equation}

Where  $A \in \mathbb{R}^{N \times N}$,  $B \in \mathbb{R}^{N}$ and $C \in \mathbb{R}^{N}$  are model parameters, and $D \in \mathbb{R}$ denotes the residual term. Here, N represents the dimension of the hidden state h(t). When the input is a continuous sequence, the above equation can model it. When the input is in a discrete form, such as an image or text, the SSMs can switch to the "zero-order hold" method to achieve discretization. Specifically, the discretization process is described by the following expression:

\begin {equation}
\begin{cases}
	$ $\overline {A}$  = EXP($\triangle A$) $   \\
	$ $\overline {B}$  = $(\triangle A)^{-1}$ (EXP($\triangle A$)- I) * $\triangle B$$   \\
	$ $\overline {C}$  = C $   
\end{cases}
\end {equation}
Where $\triangle \in \mathbb{R}^{D} $ denotes a predefined time-scale parameter that maps the continuous parameters A and B into a discrete space, and $\overline {B} \in \mathbb{R}^{D\times N}$, $\overline {C} \in \mathbb{R}^{D\times N}$ are discrete matrices. Following discretization, the computational process of the State Space Model can be represented as follows:

\begin{equation}
	\begin{cases}
		$ $h_k$ = $\overline {A}$ $h_{k-1}$ + $\overline {B}$ $X_k$  $   \\
		$ $Y_k$ = $\overline {C}$ $h_{k}$ + $\overline {D}$ $X_k$ $     
	\end{cases}
\end{equation}
Finally, the discretized SSMs can model discrete signals with linear complexity.

Selective Scan Mechanism. While SSMs are useful for efficient sequence modeling, traditional SSMs may struggle to capture complex patterns within diverse input sequences. Lacking data-dependent parameters, conventional SSMs cannot dynamically focus on or ignore specific information. To address this limitation, the Selective Scan Mechanism  for SSMs termed S6\cite{gu2023mamba} was proposed. In S6, the matrices $B \in \mathbb{R}^{L \times N}$ ,  $C \in \mathbb{R}^{L \times N}$ and the timescale parameter  $\triangle \in \mathbb{R}^{L \times D}$ are generated directly from the input data   $X \in \mathbb{R}^{L \times D}$ . This enables S6 to be contextually aware of the input, significantly enhancing its flexibility and efficiency.

Early SSMs updated their states in a fixed, input-invariant manner, making them poorly suited for processing language data requiring selective memorization. In contrast, Mamba addresses this limitation by introducing Selective SSMs, which dynamically determine which information to retain or discard based on the current input data.

\subsection{Workflow and Baseline Descriptions}

\begin{figure*}[htbp]
	\centering
	\includegraphics[width=2\columnwidth]{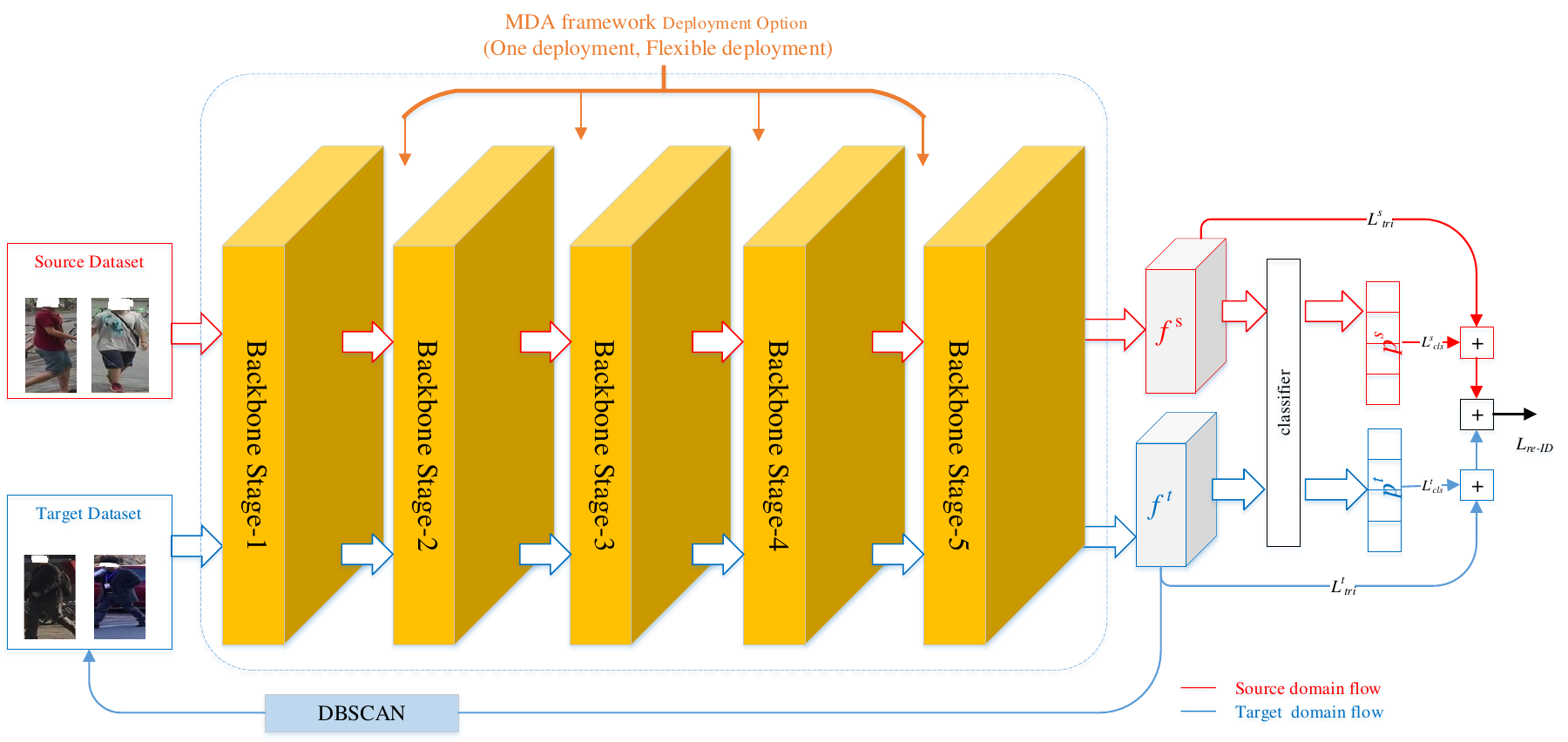}
	\caption{Workflow of domain adaptation}
	\label{fig:DApipeline}
\end{figure*}

The complete domain adaptation workflow is outlined as follows. The proposed method follows a joint training paradigm, where the training dataset consists of both labeled source samples and unlabeled target samples. During training, data are fed into the network in batches of a predetermined size. Each batch is composed of an equal proportion of labeled source samples and unlabeled target samples. The overall training pipeline comprises a backbone network, a flexibly deployable MDA framework, a clustering computation module DBSCAN\cite{Ester1996}, and the re-identification loss function $L_{ReID}$. The backbone network is divided into several stages according to feature levels. A hybrid classifier is deployed at the output of the backbone network. We apply  $L_{ReID}$ comprising the classification loss $L_{cls}$ \cite{hauptmann2016person} and triplet loss $L_{tri}$ \cite{hermans2017defense}  to both the features and predictions from source and target domains. Figure \ref{fig:DApipeline}  illustrates the domain adaptation workflow. 

\begin{equation}
	L_{ReID} = L_{cls} + L_{tri}   
\end{equation}

The deployment of the MDA framework emphasizes achieving a trade-off between computational overhead and performance gains. As a plug-and-play module, the MDA framework is deployed during both training and inference. The number of stacked MDA operators can be varied, enabling scalability of the MDA framework size. Inspired by the bottleneck structure, we implement an MDA framework comprising three composite operators. Notably, we do not concurrently deploy the MDA framework after multiple stages of the backbone network. Figure \ref{fig:MDAframework} depicts the structure of the MDA framework.

The DBSCAN\cite{Ester1996} settings are kept uniform across different datasets, although separately setting different EPS values for each dataset could yield better performance. Here, to evaluate the proposed method more precisely, this paper uniformly sets EPS to a fixed value across different target datasets.

To clarify our method, we categorize the features extracted from all stages of the backbone network into three distinct levels: (1) Low-level features, corresponding to Stage 1 and Stage 2 outputs. (2) Mid-level features, corresponding to Stage 3 outputs. (3) High-level features, corresponding to Stage 4 and Stage 5 outputs.

To validate the generalizability of the proposed method, the entire training pipeline is deployed on several baseline architectures with varied backbones.
\begin{itemize}
	\item Baseline 1: A standard baseline using a pre-trained ResNet50 \cite{he2016deep} backbone.
	\item Baseline 2: A high-performance baseline using a pre-trained IBN-ResNet50 \cite{pan2018two} backbone.
	\item Baseline 3: A baseline based on SwinTransformer \cite{9710580}.	
\end{itemize}
These three baseline types are constructed by substituting the backbone network while keeping all other components consistent.

\subsection{Introduction to MDA framework}
The design objectives of the MDA framework are briefly outlined as follows: (1) Flexibility, enabling performance improvements at different feature stages; and (2) Generality, allowing adaptation to diverse baselines. To achieve these objectives, this work progressively constructs domain adaptation capabilities, beginning with the fundamental composite operator(MDA Operator) and culminating in the complete MDA framework.

\begin{figure*}[hbp]
	\includegraphics[width=2\columnwidth]{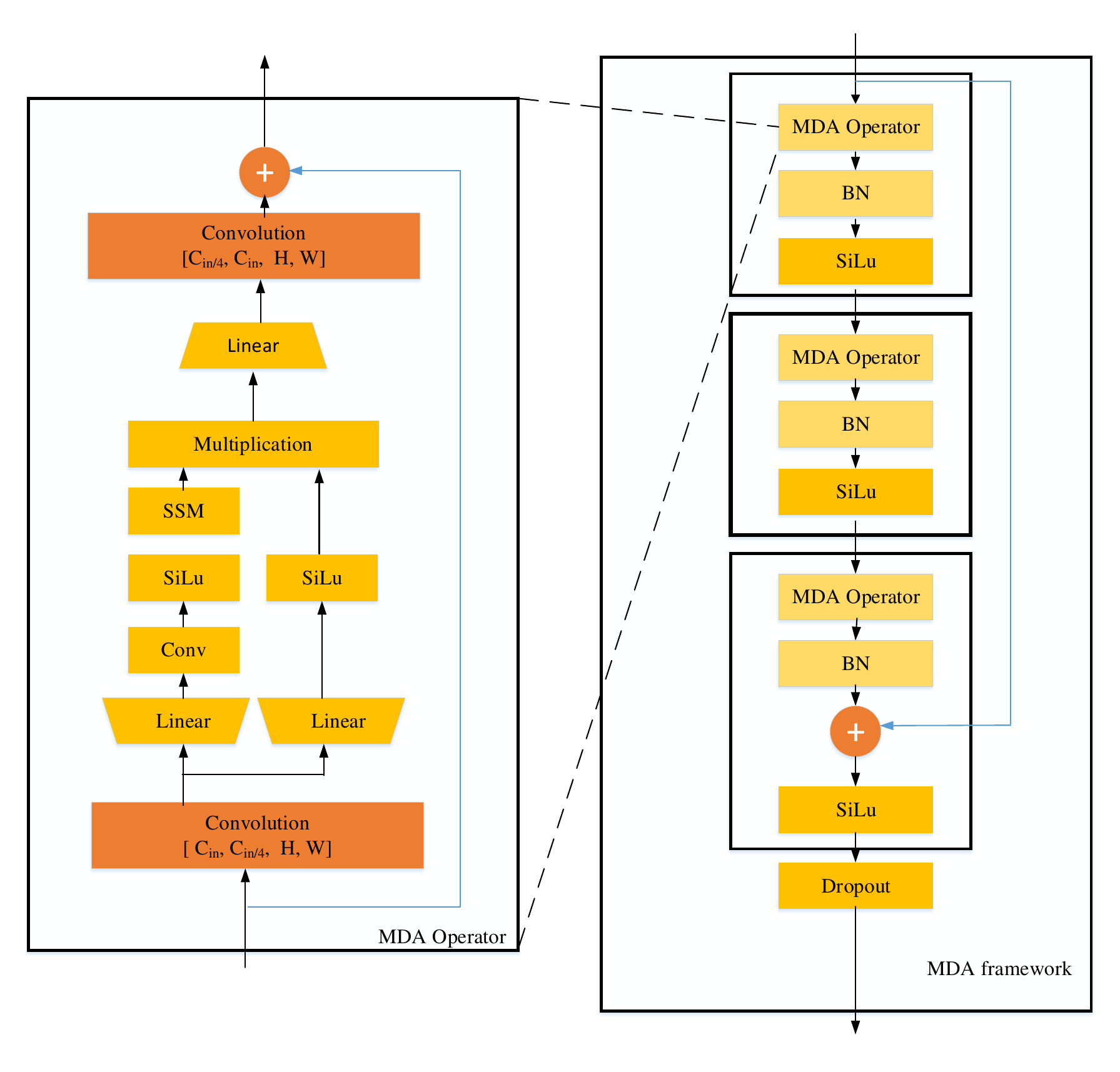}
	\caption{MDA framework and MDA Operator}
	\label{fig:MDAframework}
\end{figure*}

Serving as the fundamental operator for enhanced domain adaptation, the design of the MDA Operator inherently achieves robustness, symmetry, flexibility, consistency, and computational efficiency. (1) Robustness: A residual structure is deployed to enhance robustness. (2) Symmetry: The operator exhibits symmetry in the transformation patterns across feature map channels. (3) Flexibility: This design can avoid dependence on specific locations, thereby improving deployment flexibility. (4) Consistency: The design pursues consistent parameter settings, enabling simple configuration across multiple deployment points. (5) Computational Efficiency: A bottleneck structure is adopted to conserve computational resources. Detailed operator design is illustrated in Figure \ref{fig:MDAframework}.

Targeting multi-architecture adaptation, this paper proposes a Mamba-enhanced domain adaptation framework(MDA framework). The MDA framework employs MDA operator as its core operator, deploys batch normalization operators to perform cross-domain normalization, adopts SiLU as the activation function, and uses Dropout to alleviate overfitting. Inside the framework, the MDA operator, together with batch normalization and SiLU operators, forms a basic stacking block; at the output, Dropout that is enabled only during training is applied to mitigate overfitting. Detailed  design is illustrated in Figure \ref{fig:MDAframework}.

The scalability of the MDA framework is achieved in two ways. The first approach involves increasing the stacking number of stackable components within the framework. The second approach allows the stacking blocks themselves to be stacked to form a larger-scale framework. Scalability is a flexible option. If, at a given deployment location, a larger-scale network can improve performance, increasing the stacking number is chosen; otherwise, a single stacking instance can be adopted.

\section{Experiments}
\subsection{Datasets and Evaluation Protocols}
We validated the proposed framework using three benchmark datasets (Market-1501\cite{zheng2015person}, MSMT17\cite{wei2018person}, and CUHK03\cite{6909421}). Performance was evaluated using mean Average Precision(mAP) and Cumulative Matching Characteristics(CMC)  at $Rank-1$. During joint training, no additional supervisory information was employed apart from the ground-truth labels in the source domain and the generated pseudo-labels in the target domain.

\subsection{Implementation Details}
Based on the PyTorch framework, we designed and implemented a comprehensive domain adaptation scheme. The experimental configuration is detailed as follows:
\begin{itemize}
	\item Input Image Specifications: For baseline 1 and baseline 2, input images were resized to $256 \times128$ pixels with augmentations including random flipping, random cropping, and random erasing. Baseline 3 utilized $224 \times224$ pixel inputs.
	\item Optimizer Configuration: The Adam optimizer \cite{Kingma2014AdamAM} was deployed during training with a weight decay of 0.0005.
	\item Learning Rate Schedule: A step decay policy was applied throughout training, reducing the learning rate by a factor of 10 every 20 epochs. No warm-up strategy was implemented at the initial training stage.	
	\item Fine-tuning Protocol: Classifier parameters were fine-tuned over 50 epochs.
	\item For baselines 1 \& 2: Initial learning rate: 0.001, Batch size : 128, EPS of DBSCAN : 0.6.
	\item For baseline 3: Initial learning rate: 0.00035, Batch size : 64, EPS of DBSCAN : 0.4.
\end{itemize}


To validate the framework's transferability, we applied uniform hyperparameter settings and consistent stacking counts for MDA operator in both Baseline 1 and Baseline 3. For MDA operator, the hyperparameters include only SSM state expansion factor, local convolution width, and block expansion factor.

\subsection{Comparisons with SOTA Methods}
From the perspectives of generality, flexibility, and performance, this paper compares the proposed algorithm with existing algorithms. We selected representative unsupervised person re-identification methods from recent years to conduct performance comparisons with our approach. As shown in Table \ref{tab:table1}, when generalizing from Market-1501 to MSMT17, our method outperforms most methods from this period. When generalizing from MSMT17 to Market1501, among the methods listed in the table, some have higher baselines than our method, while others adopt optimizations on clustering settings. For a fairer comparison, our method does not apply clustering optimization tailored to different datasets, and thus its performance is lower than some methods. It should be emphasized that the methods in the table can typically only be deployed on a single architecture. In contrast, our solution can be deployed on multiple architectures with high flexibility in deployment locations.

\vspace{-0.5\baselineskip}
\begin{table}[H]
	\centering
	\tabcolsep 1pt
	\tiny{
		\caption{Comparison with the SOTA method on \\ multiple benchmarks }
		\label{tab:table1}
		
		\begin{tabular}{ccccc}
			\toprule
			\multirow{2}{*}{algorithm}  & \multicolumn{2}{c}{MSMT17 $\to$ Market}  & \multicolumn{2}{c}{Market $\to$ MSMT17}   \\ \cmidrule(rl){2-3} \cmidrule(rl){4-5} 
			& {mAP}  & R1   & {mAP}  & R1   \\ 
			\midrule
			MDJL  \cite{CHEN2023109369}    & { ---} & { ---}    & {13.4 } &  {34.3}   \\  
			\midrule
			SECRET \cite{he2022secret}    & {81.0} & {92.6} & {31.3} & 60.4   \\ 
			\midrule
			PPLR\cite{cho2022part}  & {81.5} & {92.8} & {31.4} & 61.1   \\
			\midrule
			CDCL\cite{TIAN2023111026}  & {81.9} & {93.1} & {29.5} & 59.7   \\
			\midrule
			MTNet\cite{chen2023mtnet}                              & {82.7} & 93.0 & {32.1} & 59.6 \\ 
			\midrule
			ADA framework\cite{ZHANG2026112238}                         & {83.4} & 92.8 & {23.1} & 46.0 \\ 
			\midrule
			MDA framework \textit{(Ours)}                          & \textbf{84.3} & \textbf{93.7} & \textbf{24.1} & \textbf{ 47.4} \\  
			\bottomrule
		\end{tabular}
		
	}
\end{table}

\subsection{Ablation Study}
This paper validates the effectiveness, generalizability, and consistency of the framework from the perspectives of deployment flexibility, backbone network adaptability, and parameter consistency. Experiments were conducted on multiple common datasets(CUHK03 \cite{6909421}, Market1501 \cite{zheng2015person}, MSMT17 \cite{wei2018person}),  deployed sequentially on three backbone networks, with extensive ablation studies performed using reasonable dataset combinations.

Unless otherwise specified, all subsequent experiments adhere to the following configuration: three MDA operators are deployed within the MDA framework; the MDA framework is stacked only once; and during a single experiment, the MDA framework is deployed at only one position.

\subsubsection{Experiment on Baseline 1}
To validate the performance improvement potential of the MDA framework, this paper implements a standard Baseline 1. Baseline 1 consists of the following components: backbone network ResNet-50, DBSCAN clustering, and loss function $L_{ReID}$. The performance improvement introduced by the MDA framework was tested using various dataset combinations. Table \ref{tab:table2} summarizes the performance of different dataset combinations and deployment locations under Baseline 1.

\begin{table*}[htbp]
	\centering
	\caption{Ablation experiment of MDA framework deployment on baseline 1.}
	\label{tab:table2}
	\begin{tabular}{ccccccccc}
		\toprule
		\multirow{2}{*}{Experimental setup}  & \multicolumn{2}{c}{C to M}        & \multicolumn{2}{c}{M to C}   & \multicolumn{2}{c}{M to MS}       & \multicolumn{2}{c}{MS to M}    \\ \cmidrule(rl){2-3}\cmidrule(rl){4-5}\cmidrule(rl){6-7}\cmidrule(rl){8-9}
		& {mAP}  & R1   & {mAP}  & R1  & {mAP}  & R1   & {mAP}  & R1  \\ 
		\midrule
		Baseline1                   & {61.2} & 78.9 & {6.3} & 5.5     & 13.0 & 29.1 & 71.2 & 86.5  \\  \hline
		\makecell{MDA framework   after stage 1}  & \textbf{65.0} & \textbf{81.5}   & \textbf{7.2} & \textbf{5.4}   & \textbf{14.6}   & \textbf{32.5} & \textbf{71.2} & \textbf{86.5}  \\  \hline
		\makecell{MDA framework   after stage 2} & \textbf{68.9} & \textbf{84.9} & \textbf{8.2} & \textbf{6.9}  & \textbf{15.3} & \textbf{33.7} & \textbf{73.7} & \textbf{88.5}  \\  \hline
		\makecell{MDA framework   after stage 3}  & \textbf{66.0}	& \textbf{82.5}	& \textbf{7.8}	& \textbf{6.1}  & \textbf{15.0}	& \textbf{33.0}	& \textbf{74.4} & \textbf{88.6}  \\  \hline
		\makecell{MDA framework   after stage 4} & \textbf{71.4} & \textbf{86.5} & \textbf{8.0}   & \textbf{6.4}  & \textbf{16.5}   & \textbf{35.9} & \textbf{76.2} & \textbf{89.7} \\ \hline
		\makecell{MDA framework  after stage 5} & 30.8 & 52.9 & 2.1 & 1.4  & 2.7 & 6.5 & 14.7 &32.7  \\  \hline
	\end{tabular}
	\\\vspace{1mm}\parbox{8.3cm}{Note: The characters C, M, and MS are abbreviations for CUHK03, Market1501, and MSMT17, respectively.}
\end{table*}

\subsubsection{Experiment on Baseline 2}
To assess the adaptability of the MDA framework on a high-performance baseline, this paper implements a high-performance Baseline 2. Baseline 2 consists of the following components: backbone network IBN-ResNet50, DBSCAN clustering, and loss function $L_{ReID}$. The generalization capability of the MDA framework was tested using various dataset combinations. Table \ref{tab:table3} summarizes the MDA framework's performance under Baseline 2 across different dataset combinations and deployment locations.

\begin{table*}[htbp]
	\centering
	\caption{Ablation experiment of MDA framework deployment on baseline 2.}
	\label{tab:table3}
	\begin{tabular}{ccccccccc}
		\toprule
		\multirow{2}{*}{Setting}  & \multicolumn{2}{c}{C to M}        & \multicolumn{2}{c}{M to C}   & \multicolumn{2}{c}{M to MS}       & \multicolumn{2}{c}{MS to M}    \\ \cmidrule(rl){2-3}\cmidrule(rl){4-5}\cmidrule(rl){6-7}\cmidrule(rl){8-9}
		& {mAP}  & R1   & {mAP}  & R1  & {mAP}  & R1   & {mAP}  & R1  \\ 
		\midrule
		Baseline2                   & {75.3} & 88.6 & {10.1} & 7.9     & 20.2 & 41.3 & 80.1 & 91.8  \\  \hline
		\makecell{MDA framework   after stage 1} & \textbf{78.4} & \textbf{90.2}   & \textbf{12.5} & \textbf{11.3}   & \textbf{22.8}   & \textbf{47.1} & \textbf{81.7} & \textbf{92.9}  \\  \hline
		\makecell{MDA framework   after stage 2} & \textbf{77.0} & \textbf{89.8} & \textbf{11.5} & \textbf{10.3}  & \textbf{22.8} & \textbf{46.1} & \textbf{80.6} & \textbf{91.7}  \\  \hline
		\makecell{MDA framework   after stage 3}  & \textbf{78.4}	& \textbf{89.8}	& \textbf{12.0}	& \textbf{10.6}  & \textbf{21.7}	& \textbf{44.2}	& \textbf{82.0} & \textbf{92.0}  \\  \hline
		\makecell{MDA framework   after stage 4} & \textbf{80.8} & \textbf{91.6} & \textbf{12.7}   & \textbf{10.9}  & \textbf{24.1}   & \textbf{47.4} & \textbf{84.3} & \textbf{93.7} \\ \hline
		\makecell{MDA framework   after stage 5} & {45.4} & {68.9} & {9.2} & {7.2}  & {3.3} & {7.1} & {43.2} & {67.4}  \\  \hline
	\end{tabular}
\end{table*}

\subsubsection{Experiment on Baseline 3}
To demonstrate the cross-architecture generalization capability of the MDA framework, this paper implements a Swin Transformer-based Baseline 3. Baseline 3 consists of the following components: backbone network Swin Transformer, DBSCAN clustering, and loss function  $L_{ReID}$. The cross-architecture generalization capability of the framework was tested using various dataset combinations. Table \ref{tab:table4} summarizes the MDA framework's performance under Baseline 3 across different dataset combinations and deployment locations.

\begin{table*}[htbp]
	\centering
	\caption{Ablation experiment of MDA framework deployment on baseline 3. }
	\label{tab:table4}
	\begin{tabular}{ccccccccc}
		\toprule
		\multirow{2}{*}{Experimental setup}  & \multicolumn{2}{c}{C to M}        & \multicolumn{2}{c}{M to C}   & \multicolumn{2}{c}{M to MS}       & \multicolumn{2}{c}{MS to M}    \\ \cmidrule(rl){2-3}\cmidrule(rl){4-5}\cmidrule(rl){6-7}\cmidrule(rl){8-9}
		& {mAP}  & R1   & {mAP}  & R1  & {mAP}  & R1   & {mAP}  & R1  \\ 
		\midrule
		Baseline3    & {23.9} & 48.0 & {6.9} & 6.3     & 4.8 & 14.6 & 33.2 & 59.1  \\  \hline
		\makecell{MDA framework   after stage 1} & \textbf{40.8} & \textbf{64.6}   & \textbf{8.5} & \textbf{7.4}   & \textbf{4.3}   & \textbf{10.6} & \textbf{48.3} & \textbf{70.6}  \\  \hline
		\makecell{MDA framework   after stage 2} & \textbf{41.5} & \textbf{63.2} & \textbf{8.6} & \textbf{7.5}  & \textbf{5.4} & \textbf{13.3} & \textbf{53.0} & \textbf{72.2}  \\  \hline
		\makecell{MDA framework   after stage 3}  & \textbf{43.7}	& \textbf{64.9}	& \textbf{8.3}	& \textbf{6.9}  & \textbf{4.7}	& \textbf{11.2}	& \textbf{56.7} & \textbf{76.4}  \\  \hline
		\makecell{MDA framework   after stage 4} & \textbf{36.5} & \textbf{58.9} & \textbf{5.3}   & \textbf{4.8}  & \textbf{5.7}   & \textbf{14.2} & \textbf{46.8} & \textbf{68.2} \\ \hline
		\makecell{MDA framework   after stage 5} & 40.5 & 62.6 & 9.9 & 8.4  & 4.1 & 10.3 & 57.2 & 76.9  \\  \hline
	\end{tabular}
\end{table*}

\subsubsection{Parameter Analysis of Mamba}
When migrating from Baseline 1 to Baseline 3, the hyperparameters of Mamba required zero adjustment. Deployed at all effective positions, the hyperparameters remained unchanged (local convolution width = 4, block expansion factor = 2, state expansion factor = 32).

The ease of hyperparameter tuning further validates the advantages of the framework. When migrating from Baseline 1 to Baseline 2, only minor adjustments were made to Mamba's hyperparameters. The local convolution width remained unchanged (= 4). In this paper, the state expansion factor and the block expansion factor were moderately tuned according to the spatial size and channel values of the feature maps. When deployed after Stage 1, the block expansion factor is 2 and the state expansion factor is 64. When deployed after Stages 2 and 3, the block expansion factor is 4 and the state expansion factor is 128. When deployed after Stage 4, the block expansion factor is 4 and the state expansion factor is 32.

Through a comparison of parameters and FLOPs between the MDA framework and ResNet50, the deployment of the MDA framework exhibits the following pattern: the higher the feature level, the greater the demands on FLOPs and parameters after deploying the MDA framework. Therefore, the MDA framework is most competitive when deployed at the low-level feature stage. Detailed data are presented in the table \ref{tab:table24} below.

\begin{table}[h]
	\centering
	\caption{MDA framework  VS ResNet50 stage 1/2/3/4/5 }
	\label{tab:table24}
	\begin{tabular}{ccc}
		\toprule
		Part of network & Params(M) & FLOPs(G) \\ 
		\midrule
		 ResNet50 stage 1         & 0.010      & 0.158     \\ 
		\midrule
		MDA framework after stage 1         & 0.070      & 1.048     \\ 
		\midrule
	    ResNet50 stage 2         & 0.216      & 0.895     \\ 
		\midrule
		MDA framework after stage 2          & 1.004 & 3.846     \\ 
		\midrule
		 ResNet50 stage 3         & 1.220      & 1.358     \\ 
		\midrule
		MDA framework after stage 3        & 3.930   & 3.789         \\ 
		\midrule
		 ResNet50 stage 4         & 7.098      & 1.924     \\ 
		\midrule
		MDA framework after stage 4       & 15.553   & 3.760          \\ 
		\midrule
		 ResNet50 stage 5         & 14.965     & 3.837     \\ 
		\midrule
  	   MDA framework after stage 5       & 61.875     & 14.981       \\ 
		\bottomrule
	\end{tabular}
\end{table}

\subsubsection{Analysis of Experimental Results}
Through an in-depth analysis of the experimental results obtained on the aforementioned three baselines, we draw the following conclusions:
\begin{itemize}
	\item The MDA framework demonstrates superior generalization performance.
	\item The MDA framework exhibits multi-architecture applicability.  When migrating from CNN architectures to Transformer architectures, it achieves performance improvements without re-tuning of Mamba parameters. 
	\item The MDA framework provides deployment flexibility. It demonstrated this flexibility across different architectures and can be deployed at stages corresponding to low-, medium-, and high-level features.
\end{itemize}

\subsection{Visualization}
\begin{figure*}[!h]
	\includegraphics[width=2\columnwidth]{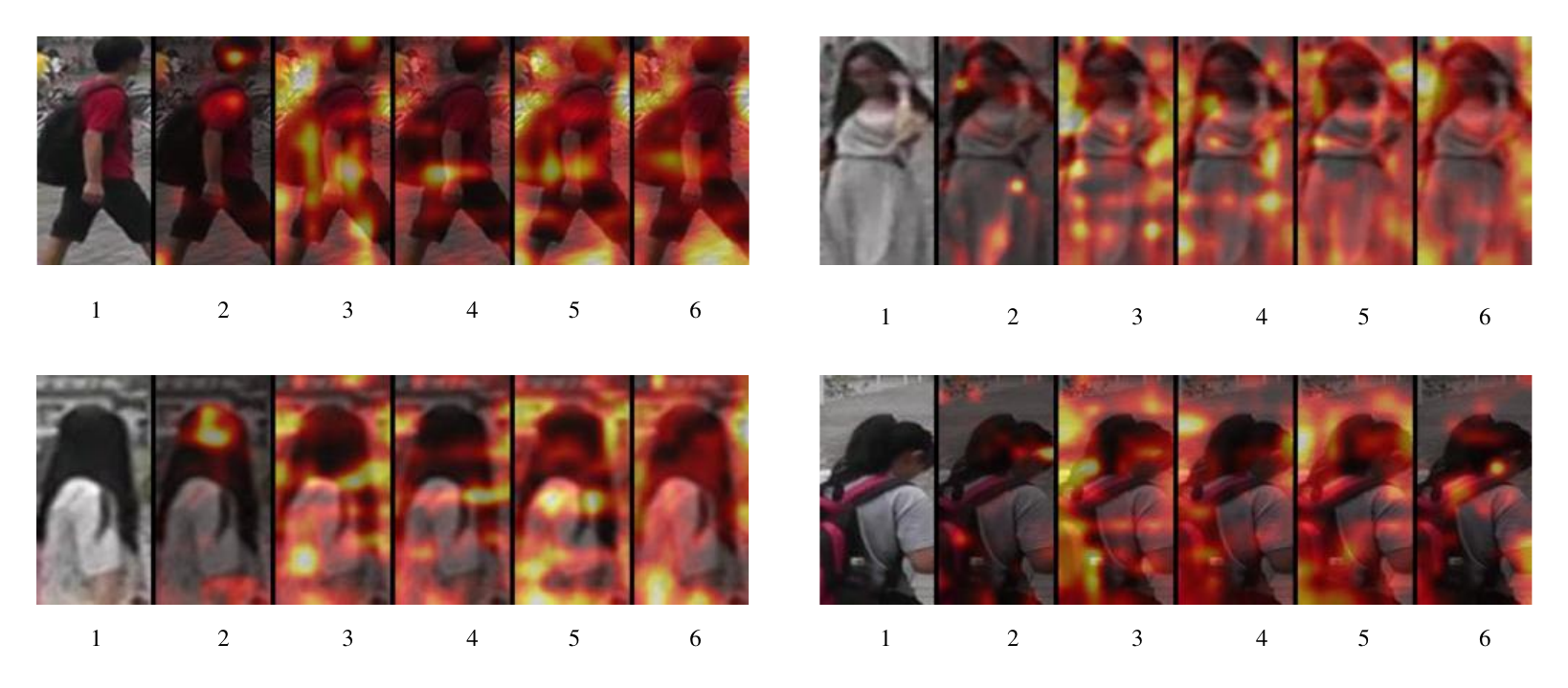}
	\caption{Visualization for MDA framework.
		Note: From left to right are figure one to six  in each group of images. 
		Figure-1 :  original image.  
		Figure-2 :  heatmap of baseline.  
		Figure-3 :  heatmap for MDA framework after stage1.  
		Figure-4:  heatmap for MDA framework after stage2.   
		Figure-5:  heatmap for MDA framework after stage3.   
		Figure-6:  heatmap for MDA framework after stage4. }
	\label{fig:heatmap}
\end{figure*}
We employed the visualization tool Torchcam\cite{torcham2020} to visualize images from the test set, using respectively the weights of Baseline 1 and the weights obtained after deploying the MDA framework at different positions within Baseline 1. The resulting visualizations were then arranged sequentially in a row. In these images, darker-colored regions (e.g., red areas) correspond to larger weights compared to lighter-colored ones. Through an in-depth analysis of the weight distributions, we conclude that, relative to the baseline output, the weight distribution exhibits pronounced selectivity after the deployment of the MDA framework. Specifically, regions relevant to identity recognition become weight-intensive, and this selectivity is introduced by the framework. Hence, the visualization results validate the effectiveness of the MDA framework. Sample visualizations are presented in Figure \ref{fig:heatmap}.

\section{Discussion}

(1) Why the framework can enhance domain adaptation. Based on the above experiments, we analyze that when an image is transformed into a long linear sequence (a large number of tokens), the framework selectively remembers or forgets information along the sequence depending on the current token. In the domain adaptation setting, this property can be converted into selectively attending to identity-relevant tokens and forgetting identity-irrelevant tokens, thereby improving recognition performance.

(2) Limitations. The built-in atomic addition operation in Mamba introduces randomness, which leads to slight performance fluctuations. This issue can be resolved through hyperparameter tuning. When Mamba-derived networks adopt this method, adjusting the network structure of the framework may be required.

(3) Why do multiple deployments and multiple stacking of the framework not further improve performance? In the experiments, the frameworks deployed at multiple positions did not form an effective synergistic effect, resulting in only minor changes in the weights of identity-relevant regions relative to identity-irrelevant regions. A similar weight modulation effect was observed with repeatedly stacked frameworks.

(4) What are the implications of this study for related fields? Constructing composite operators based on Mamba and building domain adaptation networks upon these composite operators is a feasible technical option. Since Mamba-derived networks share properties similar to those of Mamba, these derivative networks are also promising for improving domain adaptation.

\section{Conclusions}


In the field of UDA person re-identification, this paper is the first to exploit the properties of Mamba to significantly improve recognition performance. The performance improvement stems from a characteristic of the framework: The memorization of information is transformed into attention enhancement, while the forgetting of information is transformed into attention suppression. It is precisely because of this characteristic that the proposed framework can enhance recognition performance at different feature stages. The framework's adaptability to backbone networks of different architectures facilitates its application in a wider range of domains.

\bibliographystyle{cas-model2-names}

\bibliography{MDAReID}

\end{document}